\documentclass[letterpaper, 10 pt, conference, onecolumn]{IEEEtran}

\IEEEoverridecommandlockouts  

\usepackage{cite}
\usepackage{amsmath,amssymb,amsfonts}
\usepackage{algorithmic}
\usepackage{graphicx}
\usepackage{textcomp}
\usepackage{xcolor}
\usepackage{epstopdf}
\usepackage{float}
\usepackage{balance}

\title{
Jensen-Shannon Divergence Message-Passing for Rich-Text Graph Representation Learning}
\title{Jensen-Shannon Divergence Message-Passing for Rich-Text Graph Representation Learning}

\author{
\IEEEauthorblockN{Zuo Wang\textsuperscript{1*} and Ye Yuan\textsuperscript{1}}
\IEEEauthorblockA{\textsuperscript{1}{\textit{College of Computer and Information Science, Southwest University, Chongqing, China}}\\ 
wz20011013@email.swu.edu.cn\textsuperscript{*}, yuanyekl@swu.edu.cn
}
}

\begin{document}

\maketitle
\thispagestyle{empty}
\pagestyle{empty}

%%%%%%%%%%%%%%%%%%%%%%%%%%%%%%%%%%%%%%%%%%%%%%%%%%%%%%%%%%%%%%%%%%%%%%%%%%%%%%%%
\begin{abstract}
In this paper, we investigate how the widely existing contextual and structural divergence may influence the representation learning in rich-text graphs. 
To this end, we propose Jensen-Shannon Divergence Message-Passing (JSDMP), a new learning paradigm for rich-text graph representation learning.
Besides considering similarity regarding structure and text, JSDMP further captures their corresponding dissimilarity by Jensen-Shannon divergence.
Similarity and dissimilarity are then jointly used to compute new message weights among text nodes, thus enabling representations to learn with contextual and structural information from truly correlated text nodes.
With JSDMP, we propose two novel graph neural networks, namely Divergent message-passing graph convolutional network (DMPGCN) and Divergent message-passing Page-Rank graph neural networks (DMPPRG), for learning representations in rich-text graphs.
DMPGCN and DMPPRG have been extensively texted on well-established rich-text datasets and compared with several state-of-the-art baselines. 
The experimental results show that DMPGCN and DMPPRG can outperform other baselines, demonstrating the effectiveness of the proposed Jensen-Shannon Divergence Message-Passing paradigm.

\end{abstract}

% \begin{IEEEkeywords}
% Rich-text graph, Rich-text graph representation learning, Divergent Message-Passing, 
% \end{IEEEkeywords}

%%%%%%%%%%%%%%%%%%%%%%%%%%%%%%%%%%%%%%%%%%%%%%%%%%%%%%%%%%%%%%%%%%%%%%%%%%%%%%%%
\section{INTRODUCTION}
Rich-text graphs \cite{li2020survey} can effectively model the complex connections among text content, thus widely existing in the real world.
For example, scientific articles and citations can be modeled as a rich-text graph, wherein nodes and edges represent article content and citations.
User reviews on products can also formulate a rich-text graph, whose nodes and edges represent review content and social relations among users. 
Due to this ubiquitousness, representation learning in rich-text graphs has drawn much attention recently and is closely related to several important NLP applications, such as fake news detection \cite{oshikawa2020survey}, sentiment analysis \cite{wankhade2022survey}, and recommendation systems \cite{ko2022survey}.

Conventional approaches to text representation learning model text data samples (e.g., documents) as bags of words (BOW), where each word is then transformed to frequency based on TF-IDF.
With data in the form of bags of word frequency, deep neural networks, e.g., Recurrent Neural Networks (RNN) \cite{zaremba2014recurrent}, LSTM \cite{shi2015convolutional} adopted as encoders to learn representations for various tasks.
Though effective, these approaches lacks the capability of capturing the structural information (e.g., non-Euclidean structure) hidden in the text data, thus leading to degenerated performances.
To address the previously mentioned challenge faced by empirical approaches, Graph neural networks (GNNs) \cite{kipf2017semisupervised, hamilton2017inductive, wu2019simplifying, 10819283,10027700} have been adopted to learn representations from text data in the form of rich-text graphs.
Different from traditional methods such as RNN and LSTM, GNNs\cite{9357419,9159907,10.1145/3719295,9357419} is designed to handle the data with complex structures, whose representations are learned by aggregating structurally correlated data features (i.e., text features in rich-text graphs). 
GNNs have obtained state-of-the-art performances in various learning tasks from text data, especially those classification tasks \cite{li2020survey}.

\begin{figure}[t]
	\centering
\includegraphics[width=0.7\textwidth]{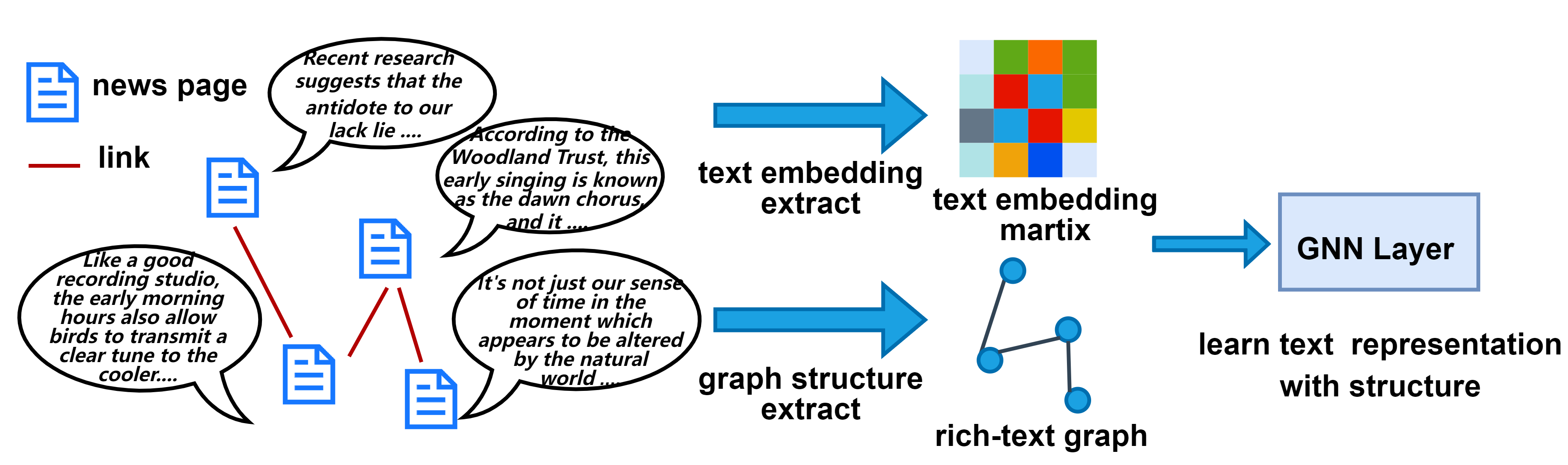}
	\caption{The general architecture of existing GNN-based approaches to text representation learning with rich-text graph data. Here, we use news pages (nodes) and corresponding references/links (edges) to construct an exemplified rich-text graph as input. 
    %General architecture of existing text representation learning methods based on GNNs. For example, given a rich-text graph, its nodes represent news pages and edges represent reference relationships among news pages.
    }
	\label{f}
\end{figure}
Existing GNN-based approaches\cite{10767725,9865020,he2021multi} to rich-text graph representation learning typically perform the learning tasks following a two-step strategy (see Fig.~\ref{f}).
First, rich-text graphs are constructed based on text embeddings extracted by sophisticated encoders (e.g., BERT) 
% yang2021bert, 
\cite{ye2020document,lu2020vgcn,piao2022sparse}, and their dependencies \cite{huang2019text,wang2022me,liu2020tensor}.
Then, a standard graph neural network, e.g., Graph convolutional network \cite{wang2022induct,li2021textgtl}, and Graph attention network \cite{veličković2018graph,CHEN2025103297,11036662,10804824,9647958}, is adopted to learn high-quality representations from the constructed rich-text graphs.
These learned representations can be used to solve various downstream classification tasks.

Compared with traditional methods, GNN-based approaches, e.g., Text-GCN \cite{yao2019graph} and BEGNN \cite{yang2021bert}, can further enhance the learning performance due to the incorporation of various GNNs\cite{10035508,11114958} that can capture the global structure hidden in the rich text.
Thus, representation learning in rich-text graphs is predominantly influenced by the capabilities of GNNs\cite{8006248,10830921}.
% Simply incorporating a more powerful GNN is expected to directly enhance the performance of representation learning in rich-text graphs.
With the emergence of massive language datasets and large language models that can provide text embeddings with more complex relationships regarding semantics, rich-text graph representation learning with GNNs\cite{10265117,9764654} is expected to play more significant roles in various real-world tasks.
Nevertheless, existing GNNs\cite{he2022neighborsworthattendingto} learn text representations from rich-text graphs mainly based on the homogeneity assumption, i.e., generating text representations by aggregating neighboring information selected according to similarity (correlation), thus lacking flexible adaptations for rich-text graph representation learning.
The contextual\cite{8766847} (e.g., dissimilarity in word frequency distributions) and structural divergence across nodes in rich-text graphs have not appropriately considered for representation learning tasks.

In this paper, we investigate how the widely existing contextual and structural divergence may influence the representation learning of rich-text graphs. 
To this end, we propose Jensen-Shannon Divergence Message-Passing (JSDMP), a new learning paradigm for rich-text graph representation learning.
Besides considering similarity regarding structure and text embeddings in the rich-text graph, JSDMP further captures their corresponding dissimilarity by Jensen-Shannon divergence \cite{nielsen2019jensen}.
Similarity and dissimilarity are then jointly used to compute new message weights among text nodes, thus enabling representations to learn with contextual and structural information from strongly correlated text nodes.
With JSDMP, we are able to build a series of graph neural networks that can learn expressive representations from rich-text graphs.
The contributions of this paper can be summarized as follows:
\begin{itemize}
    \item We present a novel learning paradigm for rich-text graph representation learning, namely Jensen-Shannon Divergence Message-Passing (JSDMP). Different from existing strategies, JSDMP considers similarity and divergence regarding word distributions and the structure of text nodes in the rich-text graph to compute the weights for aggregating neighboring information. Thus, JSDMP enables representation learning in the rich-text graph from neighbors that are genuinely correlated regarding word context and structure.
    %We propose a novel learning paradigm for rich-text graph representation learning Divergent Message-Passing (DMP), which integrates the similarity and dissimilarity in word frequency distributions between two text nodes and structural divergence.
    \item With JSDMP, we build two novel graph neural networks, including Divergent message-passing graph convolutional network (DMPGCN) and Divergent message-passing Page-Rank graph neural networks (DMPPRG) for learning expressive representations from rich-text graphs.
    %Based on DMP, we propose two novel graph neural networks, namely Divergent message-passing graph convolutional network (DMPGCN) and Divergent message-passing Page-Rank graph neural networks (DMPPRG).
    \item We conduct a series of experiments that compare the proposed DMPGCN and DMPPRG with other strong baselines on semi-supervised learning tasks from diverse real-world rich-text graphs. The experimental results demonstrate that DMPGCN and DMPPRG can outperform all compared baselines on all test datasets, showcasing the effectiveness of the proposed Divergent Message-Passing paradigm.
    %A large number of experiments have proven the effectiveness of our model in rich-text graphs, further demonstrating the importance of contextual and structural divergence for rich text classification tasks.
\end{itemize}

\section{Related works}
\subsection{Rich-text representation learning}
Rich-text representation learning has long been a classical and challenging task in natural language processing.
Early approaches to tackling this task mainly rely on shallow or hand-crafted ways to extract text features, e.g., skip-gram and bag-of-words (BoW).
Recently, deep-learning-based methods have shown their power in rich-text classification tasks.
These approaches are generally built on one or a combination of different deep learning modules.
Well-established deep learning architectures, such as Recurrent neural networks \cite{zaremba2014recurrent}, LSTM \cite{shi2015convolutional}, Transformers \cite{NIPS2017_3f5ee243} and pre-trained models \cite{devlin2018bert} have proposed to learn representations by capturing the sequential patterns among texts.
Recently, non-Euclidean and relational structures hidden in rich text data have drawn much attention.
Thus, many approaches that leverage the power of sequential encoders and well-established Graph neural networks are proposed to learn representations from rich-text data.
Typically, text embeddings are firstly learned by a well-established language encoder, e.g., BERT \cite{lin2021bertgcn}.
Then, rich-text graphs can be constructed by considering various scales of similarity measures, e.g., text level \cite{yao2019graph}, sentence level \cite{piao2022sparse}, text-document \cite{wang2022me}, and document levels \cite{ye2020document}.
Finally, a graph neural network, such as GCN \cite{liu2020tensor,lu2020vgcn,9885025} and GAT \cite{piao2022sparse} is adopted to learn representations from the constructed rich-text graphs for downstream tasks. 
It is seen that the predictive performance of these approaches to rich-text representation learning heavily rely on the learning capabilities of GNNs.
Thus, simply incorporating a more powerful GNN is expected to directly enhance the performance of representation learning in rich-text graphs.

\subsection{Graph neural networks}
% Message-Passing Graph neural networks (MPGNNs)\cite{yao2022multi} are an effective means of processing graph data. Based on the message-passing paradigm, existing GNNs can learn effective node embeddings from complex graph structures. 
Recently, Graph neural networks (GNNs) have emerged as a powerful tool for learning expressive representations for diverse downstream tasks.
Most GNNs conduct the representation learning tasks following the message-passing paradigm.
First, messages are generated from all neighboring nodes and sent to the central node. Second, the central node further aggregates the messages sent from neighboring nodes by weighting them to update the representation of the central node.
GNNs that learn representations based on the aforementioned two-step paradigm form the majority of graph representation learning, i.e., Message-passing graph neural networks \cite{kipf2017semisupervised,he2022not}.
%Message-Passing Graph Neural Networks \cite{gilmer2020message}usually consist of two steps: first, messages are generated from all neighboring nodes and sent to the central node; second, the central node further aggregates the messages sent from neighboring nodes by weighting them to update the representation of the central node.
%Current graph neural networks like GCN \cite{kipf2016semi}, GPRGNN \cite{chien2020adaptive}, Co-GNN \cite{finkelshtein2023cooperative} and others\cite{hamilton2017inductive,gasteiger2018predict,velivckovic2017graph} are all message-passing graph neural networks (MPGNNs). 
Well established GNNs including Graph convolution network (GCN) \cite{kipf2017semisupervised}, Graph Page-Rank network (GPRN) \cite{gasteiger2018predict,chien2020adaptive}, Graph attention network (GAT)
\cite{veličković2018graph,zhou2024differentiable}, and their variants \cite{zhang2020adaptive,he2021learning} all belong to Message-passing graph neural networks.
They mainly learn representations by aggregating neighboring information that is collected solely based on similarity/correlation measures.
Having investigated prevalent approaches to rich-text (graph) representation learning, we observe that there are no effective graph neural networks that can learn expressive representations by additionally considering the divergence among text data.
This motivates us to propose Divergent message-passing graph neural networks to address this gap.

%GCN\cite{kipf2016semi} relies on symmetrically normalized Laplacian matrix to aggregate messages from neighbors. GPRGNN\cite{chien2020adaptive}, which is based on APPNP\cite{gasteiger2018predict}, adaptively learns weights to sum node representations across each layer. Co-GNN\cite{finkelshtein2023cooperative} uses central node and neighboring node features to calculate the current state of central node. Then, co-GNN decides whether to send and receive messages to the neighboring nodes according to the state.

\section{Jensen-Shannon Divergence message-passing graph neural networks for rich-text representation learning}
In this section, we introduce the details of the proposed Jensen-Shannon Divergence message-passing paradigm (JSDMP).
With JSDMP, we then build several graph neural networks for representation learning in rich-text data.

\subsection{Notations}
Before introducing the proposed Jensen-Shannon Divergence message-passing paradigm, we list the necessary notations and preliminaries used in this paper.
We use $G = \left \{E, V, \mathbf F \right \}$ to represent a rich-text graph extracted from real-world text data. 
$V$ denotes the node set, representing text entities, e.g., sentences, documents, or news pages.
%$V$ denotes text entities sets, e.g, sentences, documents, or news pages. 
$E$ denotes the set of edges among text entities.
%edge set between text entities. 
$\mathbf F\in N\times D$ denotes the input word frequency matrix of the rich-text graph or text embeddings at some GNN layer. $Y_{i}$ denotes the class label to which the text entity belongs. We use $ N_{i} $ to denote the set of neighboring nodes of node $i$. $\mathbf A\in \left \{ 0,1 \right \}^{n\times n}$ represents the adjacency (with self-loops) matrix of $G$. 
$\mathbf W^{k}$, $\mathbf X$ represent the matrices of learnable weights and latent positions, respectively. The latent positions matrix $\mathbf X\in N\times C$ are randomly initialized as the one-hot matrix to represent the latent positions of nodes in the rich-text graph. 

\subsection{Jensen-Shannon Divergence message-passing paradigm}
\begin{figure*}[t]\renewcommand{\arraystretch}{1.3}
	\centering
\includegraphics[width=\columnwidth]{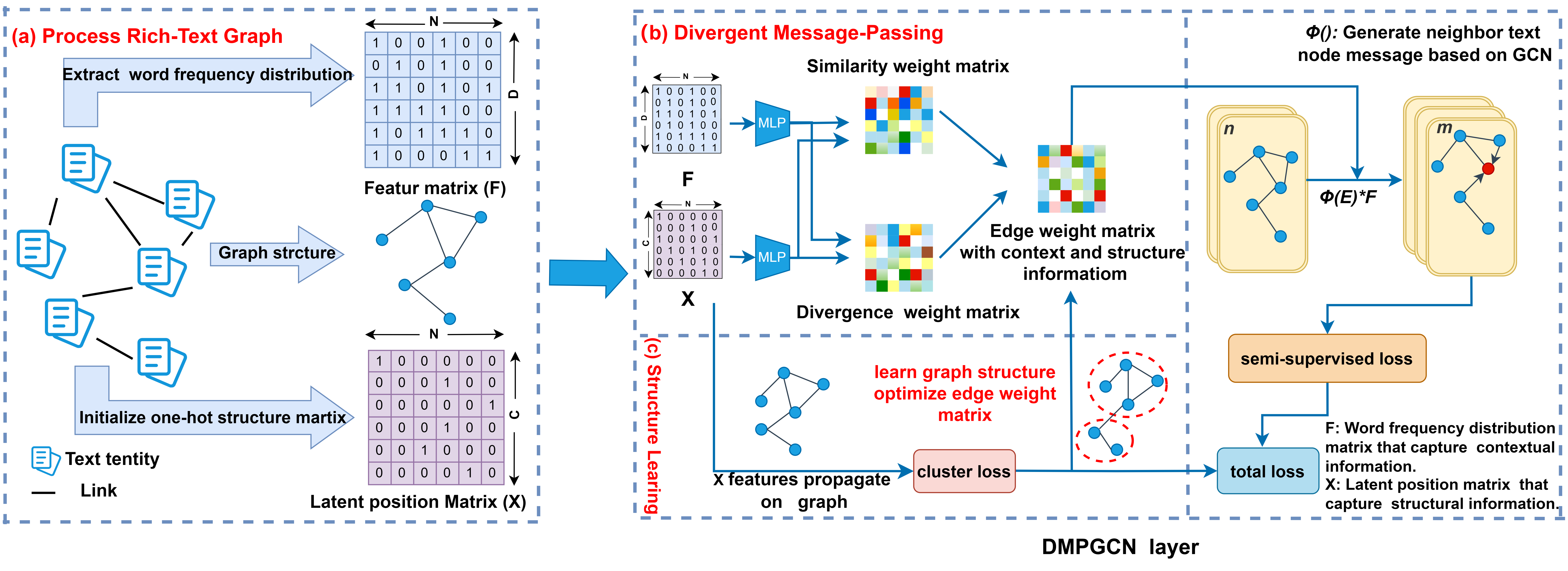}
	\caption{Overall idea of the proposed Divergent message-passing graph convolutional network (DMPGCN). $\mathbf F$ and $\mathbf X$ represent node features matrix and latent position matrix, respectively.}
	\label{frame}
\end{figure*}

% Many approaches to rich-text graph representation learning leverage GNNs \cite{wu2019simplifying,velivckovic2018deep} to explore graph topology in the rich text data. However, existing GNNs learn text representations mainly according to text similarity (correlation). In contrast, the proposed Divergent message-passing (DMP) not only considers similarity in text embeddings and their structure, but also their corresponding dissimilarity. With DMP, a graph neural network can learn expressive text representations by concentrating on truly correlated neighbors.
To achieve JSDMP, we mainly use the word frequency distribution and latent positions matrix that are processed through MLP to jointly calculate the similarity and divergence between connected text entities. 
Then, the similarity and divergence are jointly used to compute a learnable weight matrix.
%combined to obtain the learnable edge weights matrix. 
This weight matrix simultaneously considers contextual and structural similarity and divergence in rich-text graphs. 
We believe the proposed JSDMP built upon with this new weight matrix can further reinforce GNNs to learn high-quality rich-text graphs representations.
%Therefore, this new divergent message-passing paradigm can further reinforce GNNs' \cite{wang2020gcn,zhang2020adaptive} ability to learn high-quality rich-text graphs representations.
The overall framework of our method is shown in Fig.~\ref{frame}.
In what follows, we will introduce the details of the proposed Jensen-Shannon Divergence message-passing paradigm.

For each GNN layer, JSDMP firstly conducts feature mapping as the following:
%\begin{itemize}
%\item Feature Mapping: 
\begin{equation}
\begin{aligned}
&\mathbf F= \mathbf F^{k} \mathbf W^{k}_{f},\\
&\mathbf X = \mathbf X^{k} \mathbf W^{k} _{x},\\ 
\end{aligned}
\end{equation}
where $\mathbf F^{k}$ and $\mathbf X^{k}$ are input matrices of text embeddings and latent positions in k-th layer, $\mathbf W^{k}_{f}$ and $\mathbf W^{k} _{x}$ are learnable matrices in k-th layer.
With $\mathbf F$ and $\mathbf X$, we are now able to compute the weights representing similarity and divergence among nodes (text entities).
For each pair of connected nodes, we use the following strategy to compute the contextual and structural similarity:
\begin{equation}
    \mathbf S_{ij}= \mathbf A_{ij}\cdot [\mathbf a \cdot (\mathbf F_{i:} || \mathbf F_{j:})^{T} + (\mathbf X \mathbf X^{T})_{ij}] \label{3},
\end{equation}
where $\mathbf a$  is a vector of
learnable parameters, $||$ is the operator for vector concatenation.
%In the resulting $\mathbf S$ matrix, $\mathbf S_{ij}$ represents the contextual and structural similarity weights between nodes $i$ and $j$.

To capture the impact of structural and contextual divergence in the rich text data, we propose a new computational method based on Jasen-Shannon divergence.
%we use \eqref{5} to compute the JS weight of node latent positions.
Specifically, to capture the contextual divergence between each pair of nodes, we have:
\begin{equation}
\begin{aligned}
&M = \frac{1}{2}softmax (\mathbf F_{i} + \mathbf F_{j}),\\
&\mathbf D_{ij}^{'} =\frac{1}{2} \sum _{s=1}^{D} (\mathbf F_{is} \log_{}\frac{\mathbf F_{is} }{M_{s}} + \mathbf F_{js} \log_{}\frac{\mathbf F_{js} }{M_{s}}),\\
\end{aligned}
\end{equation}
where $\mathbf F_{i}$ and $\mathbf F_{j}$ represent the text embeddings of node $i$ and $j$. $D$ represents the dimension of text embeddings. 
For structural divergence, we have:
\begin{equation}
\begin{aligned}
&U = \frac{1}{2}softmax (\mathbf X_{i} + \mathbf X_{j}),\\
&\mathbf D_{ij}^{''} =\frac{1}{2} [ \sum _{c=1}^{C}  \mathbf X_{ic} \log_{}\frac{\mathbf X_{ic} }{U_{c}} + \mathbf X_{jc} \log_{}\frac{\mathbf X_{jc} }{U_{c}}] \label{5},\\
\end{aligned}
\end{equation}
where $C$ represents the dimension of each $\mathbf X_i$. 
%$\mathbf D^{''}$ is used to learn structural divergence in rich-text graphs.
With $\mathbf D^{'}$ and $\mathbf D^{''}$, we are now able to construct a new weight matrix $\mathbf D$ that simultaneously represents the contextual and structural divergence between nodes in the text graph. Accordingly, we have:
\begin{equation}
\mathbf D _{ij} = \mathbf A_{ij}\cdot  [\beta \mathbf D_{ij}^{'} +(1-\beta )\mathbf D_{ij}^{''}], 
\end{equation}
where $\beta\in (0,1)$ is a learnable parameter to balance the relative importance of contextual and structural divergence in rich-text graphs.
Given $\mathbf S$ and $\mathbf D$, JSDMP can generate a new weight matrix ($\mathbf E$) that jointly considers similarity and divergence across nodes in the text graph. And this matrix can then be used to construct different layers (i.e., JSDMP layers) to build new graph neural networks for rich-text representation learning.
For each entry of $\mathbf E$, it is computed as the following:
\begin{equation}
\mathbf E_{ij}  = \mathbf A_{ij} \cdot e^{(\mathbf S_{ij}-\gamma \mathbf D_{ij})}, 
\end{equation}
where $\gamma$ is a learnable parameter to balance the relative importance of similarity and dissimilarity.
\subsection{Divergent message-passing graph neural networks}
Based on JSDMP, we construct two new graph neural networks, namely, Divergent message-passing graph convolutional network (DMPGCN) and Divergent message-passing Page-Rank graph network (DMPPRG), where we integrate $\mathbf E$ into graph convolutional layer and graph Page-Rank layer for aggregating neighboring information for rich-text graph representation learning.

In recent, Graph convolutional neural network (GCN)
\cite{kipf2017semisupervised} has been widely adopted to 
learn representations in rich-text graphs. However, GCN does not consider how contextual and structural divergence may influence the representations learned in rich-text graphs. Therefore, we use the weight matrix ($\mathbf E$) obtained in the last subsection to construct the DMPGCN layer, which can learn high-quality rich-text graph representations.
For each DMPGCN layer, it generates the output representation ($\mathbf F_{i}^{k}$) as the following:
\begin{equation}\label{9}
\begin{aligned}
    &\phi(\mathbf E_{ij}) = \frac{\mathbf E_{ij} }{\sqrt{\sum _{n\in N_{i} } \mathbf E_{in} }\sqrt{\sum _{n\in N_{j} } \mathbf E_{jn} } }, \forall \mathbf A_{ij} = 1,\\
    %&\mathbf M_{i} = \frac{1}{\sqrt{\sum _{n\in N_{i} } \mathbf E_{in} } } \sum_{j\in N_{i}}^{}\frac{\mathbf E_{ij} }{\sqrt{\sum _{n\in N_{j} } \mathbf E_{jn} } }  \cdot \mathbf F_{j},\\
    &\mathbf F_{i}^{k} =  \sigma (\sum_{j\in N_{i}} \phi(\mathbf E_{ij})\cdot \mathbf F_{j}), \forall i,\\
    %&\mathbf M_{i}=\sum_{j\in N_{i}} \phi(\mathbf E_{ij})\cdot \mathbf F_{j},\mathbf F_{i}^{k} =  \sigma (\mathbf F_{i} + \mathbf M_{i}),\\
\end{aligned}
\end{equation}
where $\sigma(\cdot)$ is the activation function.

% Adaptive Page-Rank graph neural network (GPRGNN) \cite{chien2020adaptive} has been proven to be equivalent to polynomial graph filtering in mathematics. 
% In different graphs, GPRGNN can adaptively learn GPR parameters to fit the optimal graph filter. 
% Therefore, GPRGNN effectively improves the performance in heterogeneous graph representation learning. 

% Based on the conventional design of the GPRGNN layer, we construct the DMPPRG layer to learn expressive representations in heterogeneous rich-text graph.

\begin{figure}[t]
	\centering
\includegraphics[width=0.6\textwidth]{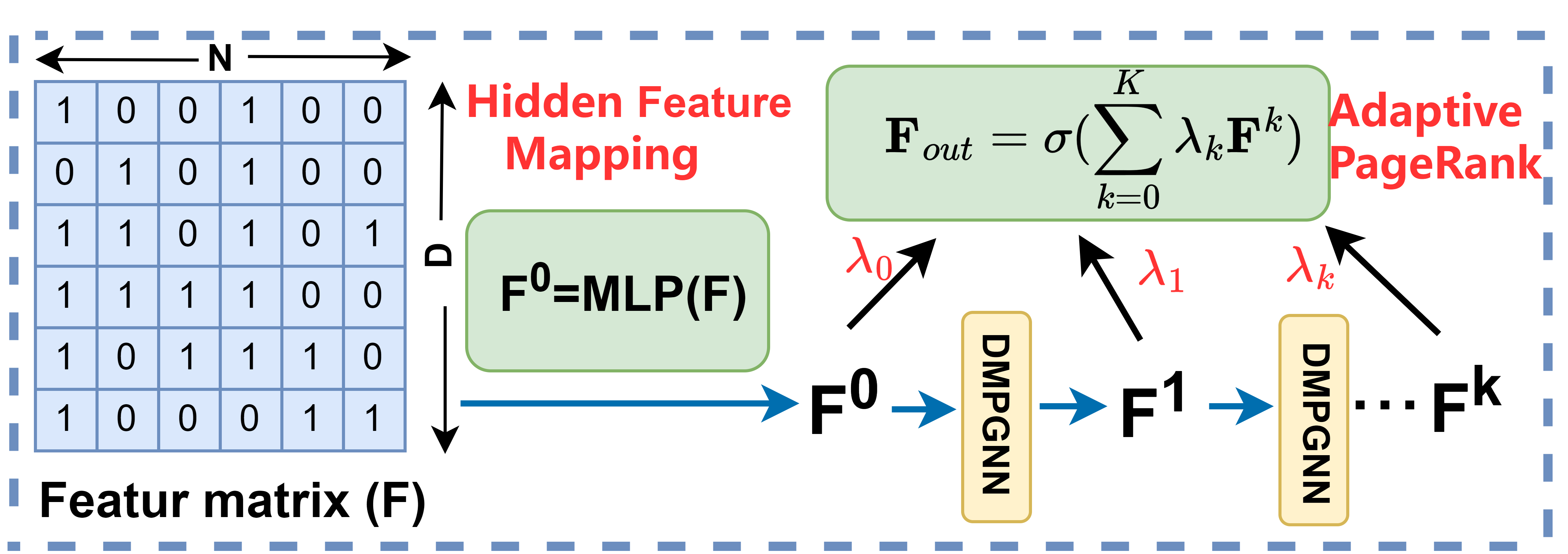}
	\caption{ 
    Overall idea of the proposed Divergent message-passing graph Page-Rank network (DMPPRG). 
    }
	\label{DMPPRG}
\end{figure}

To enhance the significance of textual representations from each layer in the final output embedding, we propose a hierarchical feature fusion mechanism based on the Adaptive PageRank Graph Neural Network (GPRGNN) \cite{chien2020adaptive}. This mechanism dynamically integrates the feature matrix $\mathbf F$ from each layer into the final output embedding by introducing learnable weighting parameters. This allows the model to autonomously determine the relative importance of features at different levels. Building upon this mechanism, we construct the DMPPRG to learn expressive representations in heterogeneous rich-text graphs.
The overall framework of DMPPRG is shown in Fig.~\ref{DMPPRG}. 
Accordingly, the output representations ($\mathbf F_{out}$) are learned as follows:
\begin{equation}
\begin{aligned}
    &\phi(\mathbf E_{ij}) = \frac{\mathbf E_{ij} }{\sqrt{\sum _{n\in N_{i} } \mathbf E_{in} }\sqrt{\sum _{n\in N_{j} } \mathbf E_{jn} } }, \forall \mathbf A_{ij} = 1,\\
    & \mathbf F_{}^{0} = MLP(\mathbf F), \mathbf F_{i}^{k} =  \sum_{j\in N_{i}} \phi(\mathbf E_{ij})\cdot \mathbf F_{j}, \forall i,\\
    & \mathbf F_{out} = \sigma(\sum_{k=0}^{K} \lambda  _{k}  \mathbf F^{k}), \\
\end{aligned}
\end{equation}
where $\lambda_{k}$ represents adaptive parameters balancing the significance of output representations learned from each layer.

\subsection{The learning objective}
% To explore the impact of structural divergence on rich-text graphs learning, $\mathbf X$ matrix is used to capture clustering information in rich-text graphs. 
The proposed DMPGCN and DMPPRG in this paper mainly consider semi-supervised learning tasks in rich-text graph data.
In addition, DMPGCN and DMPPRG leverage latent positions ($\mathbf X$) to represent the structural information of each node in the rich-text graph.
To allow $\mathbf X$ to carry more discriminated structures, we consider regularizing $\mathbf X$ based on Laplacian smoothing \cite{he2022not,zhou2024road} and orthogonal constraint\cite{zhou2024differentiable}.
Thus, we have:
\begin{equation}
\zeta _{c} = tr(\mathbf X^{T}\mathbf L\mathbf X) + \left \| \frac{1}{n} \mathbf X^{T} \mathbf X-\mathbf I \right \| _{F}^{2} ,
\end{equation}
%To enhance the impact of structural similarity and dissimilarity in rich-text graphs representation learning. DMPGCN and DMPPRG add Laplacian smoothing to the semi-supervised classification loss function. Therefore, DMPGCN and DMPPRG can learn clustering information through the latent positions matrix $\mathbf X$. 
%The auxiliary clustering information further optimizes the weight matrix $\mathbf E$. Using the $X$ matrix, we further explored the effect of structural similarity and dissimilarity in rich-text graphs representation learning.
where $\mathbf L$ is the normalized Laplacian matrix of rich-text graphs, i.e., $\mathbf L= \mathbf D^{-\frac{1}{2}}(\mathbf D-\mathbf A) \mathbf 
D^{-\frac{1}{2}}$. $\mathbf I$ denotes the identity matrix. 
%$\mathbf X$ represents latent position matrix of rich-text graphs. 
Together with the loss function for node classification ($\zeta _{semi}$), e.g., the binary cross-entropy loss, the overall loss function for DMPGCN and DMPPRG can be formulated as follows:
%$\zeta _{semi}$ represents the loss function for semi-supervised classification. The total learning objective is formulated as follows:
\begin{equation}\label{loss-fun}
\zeta  =\zeta _{semi}+\zeta_{c}.
\end{equation}
In this paper, we do not introduce any hyperparameters for $\zeta _{semi}$ and $\zeta_{c}$ as we find DMPGCN and DMPPRG perform quite well using Eq.~(\ref{loss-fun}).

\section{Experiments and analysis}\label{exp}
\begin{table}[ht]\renewcommand{\arraystretch}{1.2}
\begin{center}
	\caption{Statistics of the test datasets.}
	\label{dataset}    
	\small\centering\begin{tabular}{c| c c c c c c}
        \hline
		Dataset &$N$&$|E|$&$D$&$C$&Type\\
        \hline
	Cora&2708&5429&1433&7&\scriptsize{Scientific article}\\
        Pubmed&19717&44338&500&6&\scriptsize Scientific article\\
		Blog&5196&171743&8189&6&\scriptsize Blog page\\
		BBC&2225&9127&3121&5&\scriptsize News\\
		Guardian&6521&96279&10801&6&\scriptsize News\\
        Wikipedia&5739&524692&17311&6&\scriptsize Online encyclopedia\\
        \hline
	\end{tabular}
\end{center}
\end{table}

\renewcommand{\arraystretch}{1.2}
\begin{table*}
\caption{Performance comparisons evaluated by ACC.\\
The best result is highlighted in bold, and the second-best result is underlined.}
\label{dataset1}
\begin{center}
\small \centering \begin{tabular}{p{2cm}|p{2cm} p{2cm} p{2cm} p{2cm} p{2cm} p{2cm}}
\hline
\textbf{Method}& \textbf{Cora} & \textbf{Pubmed} & \textbf{Blog} & \textbf{BBC} & \textbf{Guardian} & \textbf{Wikipedia}\\

\hline
GCN & 0.811 \footnotesize ± 0.009	& 0.808 \footnotesize ± 0.008	& 0.511 \footnotesize± 0.003 & 0.963 \footnotesize± 0.001	 & 0.903 \footnotesize± 0.009 & 0.823 \footnotesize± 0.011		\\
GAT & 0.820 \footnotesize± 0.003 & 0.811 \footnotesize± 0.005 & 0.476 \footnotesize± 0.055	
& 0.963 \footnotesize± 0.004	 & 0.904 \footnotesize± 0.008 &0.843 \footnotesize± 0.025		\\
% HardGAT & 0.782 \footnotesize± 0.004 & 0.809 \footnotesize± 0.003 & 0.330 \footnotesize± 0.016	 & 0.955 \footnotesize± 0.004 & 0.684 \footnotesize± 0.001 &	0.462 \footnotesize± 0.062	     \\
APPNP & 0.829 \footnotesize± 0.008 & \underline{0.825} \footnotesize± 0.004 & 0.561 \footnotesize± 0.027	 & 0.964 \footnotesize± 0.004 & 0.792 \footnotesize± 0.061 &	 0.532 \footnotesize± 0.064 \\
GraphSAGE & 0.826 \footnotesize± 0.010 & 0.816 \footnotesize± 0.009 & 0.471 \footnotesize± 0.035 & 0.971 \footnotesize± 0.001 & 0.923 \footnotesize± 0.004 &  0.847 \footnotesize± 0.014\\	
SGC & 0.805 \footnotesize± 0.000 & 0.754 \footnotesize± 0.001 & 0.501 \footnotesize± 0.001		 & 0.964 \footnotesize± 0.000 & 0.913 \footnotesize± 0.000 &	 0.839 \footnotesize± 0.000	\\
 % DGI & 0.826 \footnotesize± 0.005 & 0.785 \footnotesize± 0.009 & 0.481 \footnotesize± 0.023		 & 0.967 \footnotesize± 0.003 & 0.875 \footnotesize± 0.003 &	 0.756 \footnotesize± 0.011 \\
GATv2 & \underline{0.832} \footnotesize± 0.007 & 0.803 \footnotesize± 0.013 & 0.517 \footnotesize± 0.022	 & \underline{0.972} \footnotesize± 0.003 & 0.906 \footnotesize± 0.002 &	\underline{0.869} \footnotesize± 0.006	     \\
D$^2$PT & 0.812 \footnotesize± 0.008 & 0.806 \footnotesize± 0.003 & 0.533 \footnotesize± 0.023		 & 0.969 \footnotesize± 0.007 & 0.851 \footnotesize± 0.007 &	 0.693 \footnotesize± 0.031 \\
ES-GNN & 0.813 \footnotesize± 0.007 & 0.792 \footnotesize± 0.001 & \underline{0.817} \footnotesize± 0.007 & 0.966 \footnotesize± 0.003 & \underline{0.926} \footnotesize± 0.002 & 0.860 \footnotesize± 0.014
 \\
 \hline
DMPGCN &  \textbf{0.841} \footnotesize± 0.002 & \textbf{0.840} \footnotesize± 0.001 & 0.680 \footnotesize± 0.009 &  \textbf{0.984} \footnotesize± 0.001 & \textbf{0.957} \footnotesize± 0.002 &  \textbf{0.890} \footnotesize± 0.002\\
DMPPRG &  \textbf{0.842} \footnotesize± 0.004 & \textbf{0.843} \footnotesize± 0.004 & \textbf{0.953} \footnotesize± 0.003 &  \textbf{0.981} \footnotesize± 0.001 & \textbf{0.960} \footnotesize± 0.001 &  \textbf{0.903} \footnotesize± 0.003\\
\hline
\end{tabular}
\end{center}
\end{table*}

\renewcommand{\arraystretch}{1.2}
\begin{table*}[h]
\caption{Performance comparisons evaluated by NMI.\\
The best result is highlighted in bold, and the second-best result is underlined.}
\label{dataset2}
\begin{center}
\small \centering \begin{tabular}{p{2cm}|p{2cm} p{2cm} p{2cm} p{2cm} p{2cm} p{2cm}}
\hline
\textbf{Method}& \textbf{Cora} & \textbf{Pubmed} & \textbf{Blog} & \textbf{BBC} & \textbf{Guardian} & \textbf{Wikipedia}\\
\hline
 GCN & 0.612 \footnotesize± 0.009	& 0.451 \footnotesize± 0.025	& 0.293 \footnotesize± 0.007	
 & 0.894 \footnotesize± 0.014	 & 0.767 \footnotesize± 0.005 & 0.655 \footnotesize± 0.038	\\
 GAT & 0.634 \footnotesize± 0.020 & 0.431 \footnotesize± 0.024 &  0.315 \footnotesize± 0.011
 & 0.901 \footnotesize± 0.012 & 0.775 \footnotesize± 0.007 & 0.691 \footnotesize± 0.013 \\
 % HardGAT & 0.589 \footnotesize± 0.004 & 0.431 \footnotesize± 0.013 & 0.161 \footnotesize± 0.015	 & 0.898 \footnotesize± 0.006 & 0.574 \footnotesize± 0.004 &	0.300 \footnotesize± 0.042	     \\
 APPNP & \underline{0.656} \footnotesize± 0.012 & \underline{0.467} \footnotesize± 0.005 & 0.418 \footnotesize± 0.024	 & 0.913 \footnotesize± 0.009 & 0.697 \footnotesize± 0.042 &	 0.469 \footnotesize± 0.047 \\
 GraphSAGE &  0.638 \footnotesize± 0.010 & 0.454 \footnotesize± 0.012 & 0.361 \footnotesize± 0.007 & \underline{0.929} \footnotesize± 0.009 & 0.812 \footnotesize± 0.013 &  0.716 \footnotesize± 0.009 \\	
 SGC &0.504 \footnotesize± 0.002 & 0.336 \footnotesize± 0.001 & 0.313 \footnotesize± 0.001 & 0.917 \footnotesize± 0.000 & 0.811 \footnotesize± 0.000 &	 0.682 \footnotesize± 0.000	\\
 % DGI & 0.632 \footnotesize± 0.012 & 0.399 \footnotesize± 0.003 & 0.289 \footnotesize± 0.017		 &  0.901 \footnotesize± 0.006 & 0.733 \footnotesize± 0.007 &	 0.577 \footnotesize± 0.003 \\
GATv2 & 0.646 \footnotesize± 0.011 & 0.421 \footnotesize± 0.023 & 0.354 \footnotesize± 0.007	 & 0.912 \footnotesize± 0.008 & 0.797 \footnotesize± 0.004 &	\underline{0.734} \footnotesize± 0.007	 \\
D$^2$PT & 0.631 \footnotesize± 0.008 & 0.415 \footnotesize± 0.005 & 0.393 \footnotesize± 0.017		 &  0.905 \footnotesize± 0.020 & 0.728 \footnotesize± 0.015 &	 0.489 \footnotesize± 0.035 \\
ES-GNN & 0.612 \footnotesize± 0.012 & 0.446 \footnotesize± 0.003 & \underline{0.671} \footnotesize± 0.010  & 0.891 \footnotesize± 0.002 & \underline{0.824} \footnotesize± 0.011 & 0.724 \footnotesize± 0.014			
 \\
 \hline
DMPGCN &  0.652 \footnotesize± 0.004	 & \textbf{0.500} \footnotesize± 0.003 & 0.512 \footnotesize± 0.013 &  \textbf{0.949} \footnotesize± 0.002 & \textbf{0.877} \footnotesize± 0.003 &  \textbf{0.767} \footnotesize± 0.006\\
DMPPRG &  \textbf{0.657} \footnotesize± 0.009 &  \textbf{0.499} \footnotesize± 0.010 & \textbf{0.872} \footnotesize ± 0.006 & \textbf{0.941} \footnotesize± 0.003 & \textbf{0.886} \footnotesize± 0.004 &  \textbf{0.795} \footnotesize± 0.005\\
\hline
\end{tabular}
\end{center}
\end{table*}
\subsection{Experimental setup}
In this section, we verify the effectiveness of the proposed JSDMP by comparing DMPGCN and DMPPRG with several strong baselines against the learning tasks from a number of real-world datasets of rich-text graphs. 
Additionally, we conduct a series of ablation studies to investigate the key properties of the proposed Divergent message-passing paradigm, thus revealing how contextual and structural divergence affects rich-text graph representation learning.
\paragraph{Datasets}
Six real-world rich-text graph datasets are selected to validate the effectiveness of the proposed method, including 
% Cora \cite{yang2016revisiting}, Pubmed \cite{namata2012query}, Blog \cite{huang2017label}, BBC \cite{dadgar2016novel}, Guardian \cite{dadgar2016novel}, and Wikipedia \cite{greene2014many}.
Cora, Pubmed \cite{sen2008collective}, Blog \cite{huang2017label}, BBC, Guardian, and Wikipedia \cite{greene2014many}.
Cora and Pubmed are two rich-text graphs representing the citations among scientific articles. The Blog dataset is a rich-text graph collected from the blog catalog. BBC and Guardian are two rich-text graphs collected from the news industry. Wikipedia is a rich-text graph collected from wikipedia.com. Specifically, in these rich-text graphs, node features denote word frequency distribution which extracted from textual content, e.g., scientific article title and abstract, blog content descriptions, news article and wiki page. 
% Given the diverse domains from which these rich-text graphs are collected, semi-supervised representation learning is closely related to several classification tasks. 
We have summarized the characteristics of six datasets in Table~\ref{dataset}.

\paragraph{Configurations of compared baselines}
As previously mentioned, the predictive performance of rich-text graph representation learning is mainly determined by the capabilities of GNNs.
Thus, in our experiments, we compare the proposed DMPGCN and DMPPRG with strong GNN baselines that can be used for rich-text graph representation learning, including GCN \cite{kipf2017semisupervised}, GraphSAGE \cite{hamilton2017inductive}, SGC \cite{wu2019simplifying}, GAT \cite{veličković2018graph}, APPNP\cite{gasteiger2018predict}, GATv2 \cite{brody2022how}, D$^2$PT \cite{liu2023learning}, and ES-GNN \cite{guo2024gnn}.
For fair comparisons, all compared baselines adopt a two-layer network structure to learn representations from the rich-text graphs that have been preprocessed appropriately. 
% We configure parameters using official recommended settings.
We also use official codes for all baseline GNN models and configure parameters using official recommended settings.
% Specifically, The dropout rate of GCN, APPNP, GraphSAGE, D$^2$PT, SGC , ES-GNN is set to 0.5, while the dropout rate of GAT and GATv2 is set to 0.6.
% The hidden layer dimension of GCN, GraphSAGE, GAT, D$^2$PT, GATv2, APPNP, SGC, ES-GNN is 16, 16, 256, 128, 8, 64, 8, 128, respectively.
% We adopt the Adam optimizer to train all the baselines.
On all test datasets, we run each baseline ten times to obtain the average performance.

\paragraph{Configurations of DMPGCN and DMPPRG} 
% The settings of MDPGCN and DMPPRG are similar to those of the GNN baselines. 
The dropout rates of DMPGCN and DMPPRG are 0.75 and 0.5, respectively. The learning rates of DMPGCN and DMPPRG are set to 0.01.
The hidden layer dimensions of DMPGCN and DMPPRG are set to 32 and 64, respectively. 
\paragraph{Data spilt and evaluation metrics} 
Following previous studies \cite{he2021learning,zhou2024differentiable}, we random use 140 and 60 node labels for training on Cora and Pubmed datasets.
500 and 1000 nodes are then used for validation and testing on these two datasets.
%On the scientific article classification task, the cora dataset uses 140 for training and the pubmed dataset uses 60 for training. 1,000 and 500 nodes are used for testing and validation on cora and pubmed dataset. 
For datasets Blog, BBC, Guardian, and Wikipedia, we randomly select 60\% of total samples for training, 20\% for validation, and the remaining 20\% for testing. 
% To evaluate the classification performance of all approaches on rich-text graphs, 
We adopt Accuracy (ACC) and Normalized Mutual Information (NMI) as evaluation metrics.

\subsection{Learning performance comparisons}
We selected six rich-text graph datasets from various real-world language task scenarios to evaluate the performance of all GNNs on rich text graph representation learning tasks.
% Specifically, Cora and Pubmed datasets are widely used to evaluate GNNs' performance in scientific article classification tasks. The Blog dataset is used to evaluate the performance of DMPPRG in heterogeneous rich-text graph representation learning. 
% BBC, Guardian, and Wikipedia are collected from the news industry and wikipedia.com, where very long texts widely exist, consequently causing more challenges in representation learning tasks.
The average classification performance evaluated by ACC and NMI has been listed in Tables~\ref{dataset1} and \ref{dataset2}. 

\paragraph{Classification Performance on scientific articles}
As seen in Table~\ref{dataset1} and Table~\ref{dataset2}, either DMPGCN or DMPPRG can outperform all compared baselines on Cora and Pubmed datasets. 
When evaluated by ACC, DMPGCN outperforms GCN by 3.7\% on the Cora dataset and by 4.0\% on the Pubmed dataset, respectively. DMPPRG outperforms the best baselines by 1.2\% on the Cora dataset and by 2.2\% on the Pubmed dataset, respectively.

\paragraph{Classification performance on blogs}
Classification tasks on the dataset Blog are more challenging as it is constructed following the heterogeneity assumption, i.e., edges are more likely to connect nodes (blog pages) with different labels.
For the classification task on the Blog dataset, DMPGCN and DMPPRG still perform robustly.
DMPPRG is better than any other baseline because of the hierarchical feature fusion mechanism, while DMPGCN ranks second best when compared with other GNN baselines.
When evaluated by ACC and NMI, DMPPRG outperforms the best baselines by 16.6\% and 30.0\%, respectively.

\paragraph{Classification performance on news articles and wikipedia}
% BBC, Guardian, and Wikipedia are three datasets of rich-text graphs representing the relations among long texts.
% Evaluating the performance of GNNs on these test datasets may better reveal their capabilities of learning with real-world, long articles. 
As seen in Tables~\ref{dataset1} and \ref{dataset2}, DMPGCN and DMPPRG perform the best when compared with other GNN baselines.
When evaluated by ACC, DMPGCN outperforms GCN by 2.2\% on the BBC dataset and by 6.0\% on the Guardian dataset, respectively. 
DMPPRG outperforms the best baselines by 0.1\% on the BBC dataset and by 3.7\% on the Guardian dataset, respectively.
DMPGCN outperforms GCN by 8.1\% on the Wikipedia dataset. DMPPRG outperforms the best baseline by 3.9\% on the Wikipedia dataset. 

In addition to evaluating DMPGCN and DMPPRG performance on rich-text classification, we will further reveal the advantages of Jensen-Shannon divergence message-passing by conducting ablation studies in Section \ref{as}.

% \renewcommand{\arraystretch}{1.2}
% \begin{table}
% \caption{Performance comparisons (ACC) of different variants of DMPGCN.
% %Ablation study on the components Of DMPGCN.
% }
% \label{as1}
% \begin{center}
% \small \centering \begin{tabular}{l|p{1.7cm} p{1.7cm} p{1.7cm}}
% % \small \centering \begin{tabular}{l | c c c}
% \hline
% \textbf{Components}& \textbf{Cora} & \textbf{Blog} & \textbf{Guardian}\\
% \hline
% GCN & 0.811 \footnotesize± 0.009& 0.511 \footnotesize± 0.003& 0.904 \footnotesize± 0.009\\	
% W/ structure & 0.824 \footnotesize± 0.003& 0.545 \footnotesize± 0.008	& 0.906 \footnotesize± 0.004\\
% W/ context & 0.826 \footnotesize± 0.008& 0.608 \footnotesize± 0.036	& 0.947 \footnotesize± 0.002\\
% DMPGCN & 0.836 \footnotesize± 0.003& 0.678 \footnotesize± 0.014	& 0.951 \footnotesize± 0.003\\
% \hline
% \end{tabular}
% \end{center}
% \end{table}

\begin{figure}[t]
    \centering
    \includegraphics[width=0.64\columnwidth]{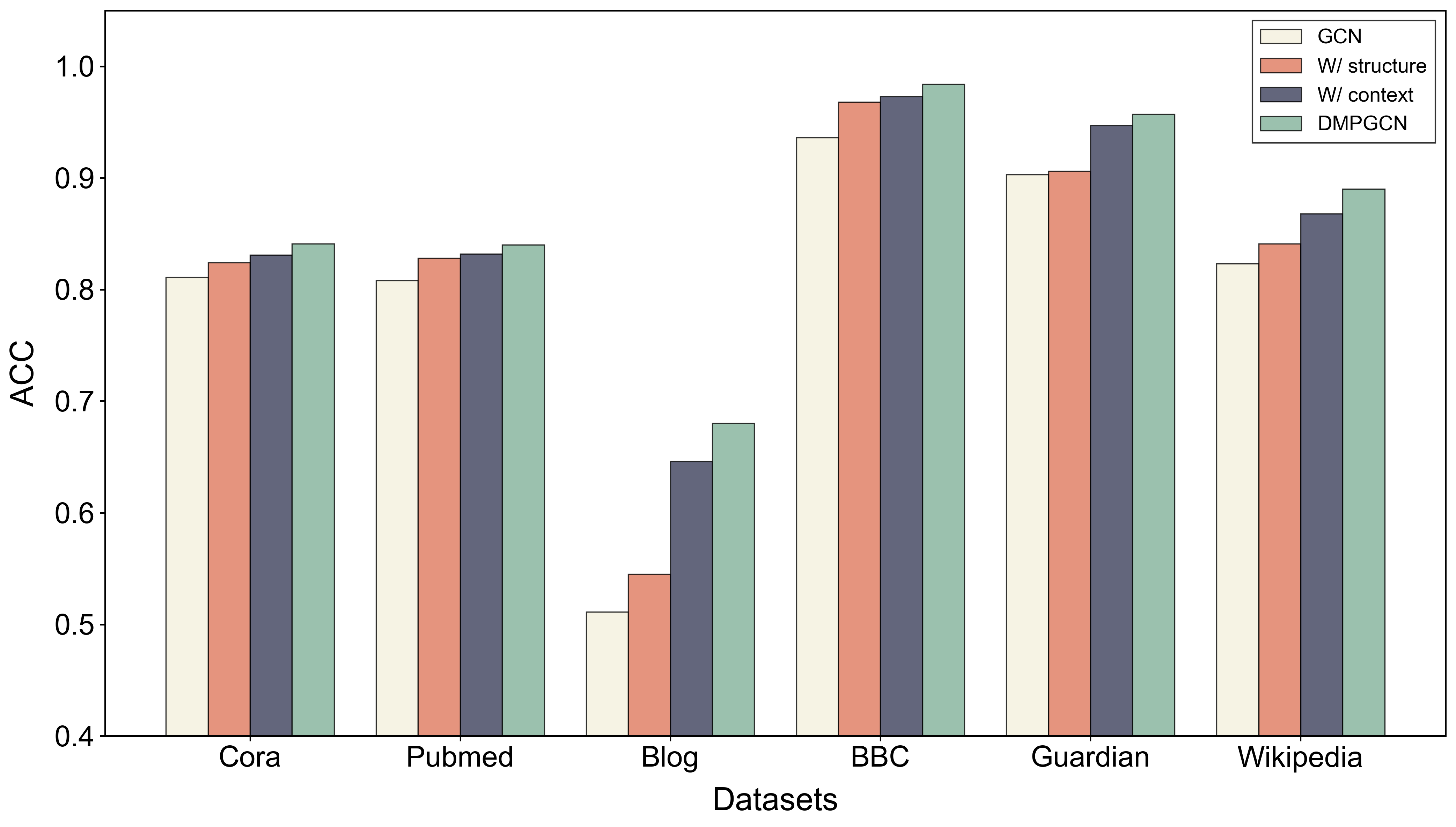} % 稍小于单栏宽度
    \caption{Performance comparisons (ACC) of different variants of DMPGCN.}
    \label{ACC}
\end{figure}

\subsection{Ablation studies} \label{as}
To reveal how contextual and structural divergence may influence representation learning in rich-text graphs, we separately investigate the effects of contextual and structural divergence in Jensen-Shannon divergence message-passing. With DMPGCN, we add structural (W/ structure) and contextual (W/ context) similarity and divergence to evaluate the performance of rich-text graph representation learning. 
The ablation results are illustrated in Figure~\ref{ACC}.
% As seen in these tables, the performance of rich-text representation learning is evidently improved with the incorporation of structural and contextual divergence into GCN.
% For example, ACC is improved by 0.3\% and 5.3\% on the Guardian dataset, 2.2\% and 7.8\% on the Wikipedia dataset, and 6.7\% and 32.7\% on the Blog dataset when structural divergence (DMPGCN w/ structure) and both contextual and structural divergence (DMPGCN) are considered by the proposed DMP (see Table \ref{as1}). 
% Similarly, NMI is improved by 2.6\% and 14.3\% on the Guardian dataset, 6.4\%, and 16.0\% on the Wikipedia dataset when considering only structural divergence and both contextual and structural divergence (see Table \ref{as2}).
From the ablation results, we observe that contextual and structural divergence brings non-trivial performance gain on most of the rich-text graphs datasets.

% \subsection{Visualization of latent position representations}
% In this part, we discuss that Divergent Message-Passing can learn text representations with structural similarity and divergence. To provide a more intuitive visualization, we utilize t-SNE to project the latent position representations X into a 2-dimensional space. As shown in Figure~\ref{x},~\ref{y}, structurally similar nodes are close together, and structurally dissimilar nodes are far away.

% Latent position matrix X can learn representations with more structural similarity and divergence through DMPGCN layer.

\section{Conclusions}
In this paper, we have proposed the Jensen-Shannon Divergence Message-Passing (JSDMP) paradigm, leveraging which we can build novel graph neural networks for rich-text graph representation learning.
%In this paper, we propose Divergent message-passing (DMP) paradigm, which is a new GNNs design strategy for rich-text graph representation learning. 
Besides considering similarity regarding structure and text embeddings, JSDMP further considers the widely existing contextual and structural divergence in rich-text graphs. Building upon this foundation, we construct two novel graph neural networks, DMPGCN and DMPPRG, which can learn high-quality representations from strongly correlated neighboring nodes in the rich-text graph.
%Through the fusion of contextual and structural similarity and dissimilarity in rich-text graphs, DMPGCN and DMPPRG can learn high-quality text representations from strongly correlated neighboring nodes. 
DMPGCN and DMPPRG have been compared with several strong GNNs for rich-text graph representation learning on challenging datasets collected from diverse scenarios.
The obtained results show that the two proposed GNNs can perform the best on most test datasets, indicating the effectiveness of the proposed JSDMP.
%Through extensive experiments, we validated the effectiveness of our model in rich-text graphs representation learning. 
In the future, we will focus on integrating JSDMP with large models to solve more challenging natural language processing tasks, such as sentiment analysis and language modeling.

\bibliographystyle{IEEEtran}
\bibliography{ref}

@article{hamilton2017inductive,
  title={Inductive representation learning on large graphs},
  author={Hamilton, Will and Ying, Zhitao and Leskovec, Jure},
  journal={Advances in neural information processing systems},
  volume={30},
  year={2017}
}

@inproceedings{NIPS2017_3f5ee243,
 author = {Vaswani, Ashish and Shazeer, Noam and Parmar, Niki and Uszkoreit, Jakob and Jones, Llion and Gomez, Aidan N and Kaiser, \L ukasz and Polosukhin, Illia},
 booktitle = {Advances in Neural Information Processing Systems},
 publisher = {Curran Associates, Inc.},
 title = {Attention is All you Need},
 volume = {30},
 year = {2017}
}

@article{ko2022survey,
  title={A survey of recommendation systems: recommendation models, techniques, and application fields},
  author={Ko, Hyeyoung and Lee, Suyeon and Park, Yoonseo and Choi, Anna},
  journal={Electronics},
  volume={11},
  number={1},
  pages={141},
  year={2022},
  publisher={MDPI}
}

@inproceedings{gasteiger2018predict,
  title={Predict then Propagate: Graph Neural Networks meet Personalized PageRank},
  author={Gasteiger, Johannes and Bojchevski, Aleksandar and G{\"u}nnemann, Stephan},
  booktitle={International Conference on Learning Representations},
  year={2018}
}

@article{chien2020adaptive,
  title={Adaptive universal generalized pagerank graph neural network},
  author={Chien, Eli and Peng, Jianhao and Li, Pan and Milenkovic, Olgica},
  journal={arXiv preprint arXiv:2006.07988},
  year={2020}
}

@inproceedings{yao2019graph,
  title={Graph convolutional networks for text classification},
  author={Yao, Liang and Mao, Chengsheng and Luo, Yuan},
  booktitle={Proceedings of the AAAI conference on artificial intelligence},
  volume={33},
  number={01},
  pages={7370--7377},
  year={2019}
}

@article{nielsen2019jensen,
  title={On the Jensen--Shannon symmetrization of distances relying on abstract means},
  author={Nielsen, Frank},
  journal={Entropy},
  volume={21},
  number={5},
  pages={485},
  year={2019},
  publisher={MDPI}
}

@inproceedings{wu2019simplifying,
  title={Simplifying graph convolutional networks},
  author={Wu, Felix and Souza, Amauri and Zhang, Tianyi and Fifty, Christopher and Yu, Tao and Weinberger, Kilian},
  booktitle={International conference on machine learning},
  pages={6861--6871},
  year={2019},
  organization={PMLR}
}

@article{he2021learning,
  title={Learning conjoint attentions for graph neural nets},
  author={He, Tiantian and Ong, Yew Soon and Bai, Lu},
  journal={Advances in Neural Information Processing Systems},
  volume={34},
  pages={2641--2653},
  year={2021}
}

@article{zhou2024differentiable,
  title={Differentiable Clustering for Graph Attention},
  author={Zhou, Haicang and He, Tiantian and Ong, Yew-Soon and Cong, Gao and Chen, Quan},
  journal={IEEE Transactions on Knowledge and Data Engineering},
  year={2024},
  publisher={IEEE}
}

@inproceedings{zhang2020adaptive,
  title={Adaptive structural fingerprints for graph attention networks},
  author={Zhang, Kai and Zhu, Yaokang and Wang, Jun and Zhang, Jie},
  booktitle={International Conference on Learning Representations},
  year={2020}
}

@inproceedings{huang2019text,
  title={Text Level Graph Neural Network for Text Classification},
  author={Huang, Lianzhe and Ma, Dehong and Li, Sujian and Zhang, Xiaodong and Wang, Houfeng},
  booktitle={Proceedings of the 2019 Conference on Empirical Methods in Natural Language Processing and the 9th International Joint Conference on Natural Language Processing (EMNLP-IJCNLP)},
  pages={3444--3450},
  year={2019}
}

@article{guo2024gnn,
  title={ES-GNN: Generalizing graph neural networks beyond homophily with edge splitting},
  author={Guo, Jingwei and Huang, Kaizhu and Zhang, Rui and Yi, Xinping},
  journal={IEEE Transactions on Pattern Analysis and Machine Intelligence},
  year={2024},
  publisher={IEEE}
}

@inproceedings{lin2021bertgcn,
  title={BertGCN: Transductive Text Classification by Combining GNN and BERT},
  author={Lin, Yuxiao and Meng, Yuxian and Sun, Xiaofei and Han, Qinghong and Kuang, Kun and Li, Jiwei and Wu, Fei},
  booktitle={Findings of the Association for Computational Linguistics: ACL-IJCNLP},
  pages={1456--1462},
  year={2021}
}

@article{li2020survey,
  title={A survey on text classification: From shallow to deep learning},
  author={Li, Qian and Peng, Hao and Li, Jianxin and Xia, Congying and Yang, Renyu and Sun, Lichao and Yu, Philip S and He, Lifang},
  journal={arXiv preprint arXiv:2008.00364},
  year={2020}
}

@article{yang2021bert,
  title={Bert-enhanced text graph neural network for classification},
  author={Yang, Yiping and Cui, Xiaohui},
  journal={Entropy},
  volume={23},
  number={11},
  pages={1536},
  year={2021},
  publisher={MDPI}
}

@inproceedings{oshikawa2020survey,
  title={A Survey on Natural Language Processing for Fake News Detection},
  author={Oshikawa, Ray and Qian, Jing and Wang, William Yang},
  booktitle={Proceedings of the Twelfth Language Resources and Evaluation Conference},
  pages={6086--6093},
  year={2020}
}

@article{zaremba2014recurrent,
  title={Recurrent neural network regularization},
  author={Zaremba, Wojciech},
  journal={arXiv preprint arXiv:1409.2329},
  year={2014}
}

@article{shi2015convolutional,
  title={Convolutional LSTM network: A machine learning approach for precipitation nowcasting},
  author={Shi, Xingjian and Chen, Zhourong and Wang, Hao and Yeung, Dit-Yan and Wong, Wai-Kin and Woo, Wang-chun},
  journal={Advances in neural information processing systems},
  volume={28},
  year={2015}
}

@article{wankhade2022survey,
  title={A survey on sentiment analysis methods, applications, and challenges},
  author={Wankhade, Mayur and Rao, Annavarapu Chandra Sekhara and Kulkarni, Chaitanya},
  journal={Artificial Intelligence Review},
  volume={55},
  number={7},
  pages={5731--5780},
  year={2022},
  publisher={Springer}
}

@inproceedings{wang2022induct,
  title={Induct-gcn: Inductive graph convolutional networks for text classification},
  author={Wang, Kunze and Han, Soyeon Caren and Poon, Josiah},
  booktitle={26th International Conference on Pattern Recognition (ICPR)},
  pages={1243--1249},
  year={2022}
}

@inproceedings{wang2022me,
  title={ME-GCN: Multi-dimensional Edge-Embedded Graph Convolutional Networks for Semi-supervised Text Classification},
  author={Wang, Kunze and Han, Caren and Long, Siqu and Poon, Josiah},
  booktitle={ICLR 2022 Workshop on Deep Learning on Graphs for Natural Language Processing},
  year={2022}
}

@inproceedings{li2021textgtl,
  title={TextGTL: Graph-based Transductive Learning for Semi-supervised Text Classification via Structure-Sensitive Interpolation},
  author={Li, Chen and Peng, Xutan and Peng, Hao and Li, Jianxin and Wang, Lihong},
  booktitle={IJCAI},
  pages={2680--2686},
  year={2021}
}

@incollection{ye2020document,
  title={Document and word representations generated by graph convolutional network and bert for short text classification},
  author={Ye, Zhihao and Jiang, Gongyao and Liu, Ye and Li, Zhiyong and Yuan, Jin},
  booktitle={ECAI},
  pages={2275--2281},
  year={2020}
}

@inproceedings{lu2020vgcn,
  title={VGCN-BERT: augmenting BERT with graph embedding for text classification},
  author={Lu, Zhibin and Du, Pan and Nie, Jian-Yun},
  booktitle={Advances in Information Retrieval: 42nd European Conference on IR Research},
  pages={369--382},
  year={2020}
}

@inproceedings{piao2022sparse,
  title={Sparse structure learning via graph neural networks for inductive document classification},
  author={Piao, Yinhua and Lee, Sangseon and Lee, Dohoon and Kim, Sun},
  booktitle={Proceedings of the AAAI conference on artificial intelligence},
  volume={36},
  number={10},
  pages={11165--11173},
  year={2022}
}

@inproceedings{liu2020tensor,
  title={Tensor graph convolutional networks for text classification},
  author={Liu, Xien and You, Xinxin and Zhang, Xiao and Wu, Ji and Lv, Ping},
  booktitle={Proceedings of the AAAI conference on artificial intelligence},
  volume={34},
  number={05},
  pages={8409--8416},
  year={2020}
}

@article{devlin2018bert,
  title={Bert: Pre-training of deep bidirectional transformers for language understanding},
  author={Devlin, Jacob},
  journal={arXiv preprint arXiv:1810.04805},
  year={2018}
}

@inproceedings{greene2014many,
  title={How Many Topics? Stability Analysis for Topic Models},
  author={Greene, Derek and O’Callaghan, Derek and Cunningham, P{\'a}draig},
  booktitle={Joint European Conference on Machine Learning and Knowledge Discovery in Databases},
  pages={498--513},
  year={2014}
}

@inproceedings{huang2017label,
  title={Label informed attributed network embedding},
  author={Huang, Xiao and Li, Jundong and Hu, Xia},
  booktitle={Proceedings of the tenth ACM international conference on web search and data mining},
  pages={731--739},
  year={2017}
}

@article{he2022not,
  title={Not all neighbors are worth attending to: Graph selective attention networks for semi-supervised learning},
  author={He, Tiantian and Zhou, Haicang and Ong, Yew-Soon and Cong, Gao},
  journal={arXiv preprint arXiv:2210.07715},
  year={2022}
}

@inproceedings{liu2023learning,
  title={Learning strong graph neural networks with weak information},
  author={Liu, Yixin and Ding, Kaize and Wang, Jianling and Lee, Vincent and Liu, Huan and Pan, Shirui},
  booktitle={Proceedings of the 29th ACM SIGKDD Conference on Knowledge Discovery and Data Mining},
  pages={1559--1571},
  year={2023}
}

@inproceedings{kipf2017semisupervised,
title={Semi-Supervised Classification with Graph Convolutional Networks},
author={Thomas N. Kipf and Max Welling},
booktitle={International Conference on Learning Representations},
year={2017}
}

@inproceedings{
veličković2018graph,
title={Graph Attention Networks},
author={Petar Veličković and Guillem Cucurull and Arantxa Casanova and Adriana Romero and Pietro Liò and Yoshua Bengio},
booktitle={International Conference on Learning Representations},
year={2018},
}

@inproceedings{brody2022how,
title={How Attentive are Graph Attention Networks? },
author={Shaked Brody and Uri Alon and Eran Yahav},
booktitle={International Conference on Learning Representations},
year={2022},
}

@article{sen2008collective,
  title={Collective classification in network data},
  author={Sen, Prithviraj and Namata, Galileo and Bilgic, Mustafa and Getoor, Lise and Galligher, Brian and Eliassi-Rad, Tina},
  journal={AI magazine},
  volume={29},
  number={3},
  pages={93--93},
  year={2008}
}

@article{zhou2024road,
  title={Road network representation learning with the third law of geography},
  author={Zhou, Haicang and Huang, Weiming and Chen, Yile and He, Tiantian and Cong, Gao and Ong, Yew Soon},
  journal={Advances in Neural Information Processing Systems},
  volume={37},
  pages={11789--11813},
  year={2024}
}

@ARTICLE{10819283,
  author={Bi, Fanghui and He, Tiantian and Ong, Yew-Soon and Luo, Xin},
  journal={IEEE Transactions on Knowledge and Data Engineering}, 
  title={Graph Linear Convolution Pooling for Learning in Incomplete High-Dimensional Data}, 
  year={2025},
  volume={37},
  number={4},
  pages={1838-1852}}

@ARTICLE{11114958,
  author={Bi, Fanghui and He, Tiantian and Ong, Yew-Soon and Luo, Xin},
  journal={IEEE Transactions on Neural Networks and Learning Systems}, 
  title={Discovering Spatiotemporal–Individual Coupled Features From Nonstandard Tensors—A Novel Dynamic Graph Mixer Approach}, 
  year={2025},
  volume={36},
  number={11},
  pages={19834-19848},
  doi={10.1109/TNNLS.2025.3592692}}

@INPROCEEDINGS{10027700,
  author={Bi, Fanghui and He, Tiantian and Luo, Xin},
  booktitle={2022 IEEE International Conference on Data Mining (ICDM)}, 
  title={A Two-Stream Light Graph Convolution Network-based Latent Factor Model for Accurate Cloud Service QoS Estimation}, 
  year={2022},
  pages={855-860},
  doi={10.1109/ICDM54844.2022.00097}}

@ARTICLE{10035508,
  author={Bi, Fanghui and He, Tiantian and Xie, Yuetong and Luo, Xin},
  journal={IEEE Transactions on Services Computing}, 
  title={Two-Stream Graph Convolutional Network-Incorporated Latent Feature Analysis}, 
  year={2023},
  volume={16},
  number={4},
  pages={3027-3042},
  doi={10.1109/TSC.2023.3241659}}

@ARTICLE{10265117,
  author={Bi, Fanghui and He, Tiantian and Luo, Xin},
  journal={IEEE Transactions on Services Computing}, 
  title={A Fast Nonnegative Autoencoder-Based Approach to Latent Feature Analysis on High-Dimensional and Incomplete Data}, 
  year={2024},
  volume={17},
  number={3},
  pages={733-746},
  doi={10.1109/TSC.2023.3319713}}

@ARTICLE{9764654,
  author={Xie, Yu and Liang, Yanfeng and Gong, Maoguo and Qin, A. K. and Ong, Yew-Soon and He, Tiantian},
  journal={IEEE Transactions on Cybernetics}, 
  title={Semisupervised Graph Neural Networks for Graph Classification}, 
  year={2023},
  volume={53},
  number={10},
  pages={6222-6235},
  doi={10.1109/TCYB.2022.3164696}}

@INPROCEEDINGS{10830921,
  author={Bi, Fanghui and He, Tiantian},
  booktitle={2024 IEEE International Conference on Systems, Man, and Cybernetics (SMC)}, 
  title={SCG: A Novel Spatiotemporal Coupling Graph Convolutional Network-Incorporated Approach for Dynamic QoS Estimation}, 
  year={2024},
  volume={},
  number={},
  pages={635-640},
  doi={10.1109/SMC54092.2024.10830921}}

@ARTICLE{10767725,
  author={Yao, Lizhong and Zhang, Yu and Wang, Ling and Li, Rui and He, Tiantian},
  journal={IEEE Transactions on Industrial Informatics}, 
  title={Natural Gas Pipeline Leak Detection Based on Dual Feature Drift in Acoustic Signals}, 
  year={2025},
  volume={21},
  number={2},
  pages={1950-1959},
  doi={10.1109/TII.2024.3495787}}

@ARTICLE{9865020,
  author={Liu, Zhigang and Yuan, Guangxiao and Luo, Xin},
  journal={IEEE/CAA Journal of Automatica Sinica}, 
  title={Symmetry and Nonnegativity-Constrained Matrix Factorization for Community Detection}, 
  year={2022},
  volume={9},
  number={9},
  pages={1691-1693},
  doi={10.1109/JAS.2022.105794}}

@article{he2021multi, title={Multi-source propagation aware network clustering}, author={He, Tiantian and Ong, Yew-Soon and Hu, Pengwei}, journal={Neurocomputing}, volume={453}, pages={119--130}, year={2021}, publisher={Elsevier} }

@ARTICLE{8766847,
  author={He, Tiantian and Liu, Yang and Ko, Tobey H. and Chan, Keith C. C. and Ong, Yew-Soon},
  journal={IEEE Transactions on Cybernetics}, 
  title={Contextual Correlation Preserving Multiview Featured Graph Clustering}, 
  year={2020},
  volume={50},
  number={10},
  pages={4318-4331},
  doi={10.1109/TCYB.2019.2926431}}

@article{CHEN2025103297,
title = {Enhancing graph convolutional networks with an efficient k-hop neighborhood approach},
journal = {Information Fusion},
volume = {124},
pages = {103297},
year = {2025},
issn = {1566-2535},
doi = {https://doi.org/10.1016/j.inffus.2025.103297},
url = {https://www.sciencedirect.com/science/article/pii/S1566253525003707},
author = {Jiufang Chen and Xin Luo and Ye Yuan and Zidong Wang},
}

@ARTICLE{11036662,
  author={Yuan, Ye and Lu, Siyang and Luo, Xin},
  journal={IEEE/CAA Journal of Automatica Sinica}, 
  title={A Proportional Integral Controller-Enhanced Non-Negative Latent Factor Analysis Model}, 
  year={2025},
  volume={12},
  number={6},
  pages={1246-1259},
  doi={10.1109/JAS.2024.125055}}

@ARTICLE{10804824,
  author={Yuan, Ye and Wang, Ying and Luo, Xin},
  journal={IEEE Transactions on Neural Networks and Learning Systems}, 
  title={A Node-Collaboration-Informed Graph Convolutional Network for Highly Accurate Representation to Undirected Weighted Graph}, 
  year={2025},
  volume={36},
  number={6},
  pages={11507-11519},
  doi={10.1109/TNNLS.2024.3514652}}

@ARTICLE{9647958,
  author={Luo, Xin and Wu, Hao and Wang, Zhi and Wang, Jianjun and Meng, Deyu},
  journal={IEEE Transactions on Pattern Analysis and Machine Intelligence}, 
  title={A Novel Approach to Large-Scale Dynamically Weighted Directed Network Representation}, 
  year={2022},
  volume={44},
  number={12},
  pages={9756-9773},
  doi={10.1109/TPAMI.2021.3132503}}

@misc{he2022neighborsworthattendingto,
      title={Not All Neighbors Are Worth Attending to: Graph Selective Attention Networks for Semi-supervised Learning}, 
      author={Tiantian He and Haicang Zhou and Yew-Soon Ong and Gao Cong},
      year={2022},
      eprint={2210.07715},
      archivePrefix={arXiv},
      primaryClass={cs.LG},
      url={https://arxiv.org/abs/2210.07715}, 
}

@article{10.1145/3719295,
author = {Xu, Xiuqin and Lin, Mingwei and Xu, Zeshui and Luo, Xin},
title = {Attention-Mechanism-Based Neural Latent-Factorization-of-Tensors Model},
year = {2025},
issue_date = {May 2025},
publisher = {Association for Computing Machinery},
address = {New York, NY, USA},
volume = {19},
number = {4},
issn = {1556-4681},
url = {https://doi.org/10.1145/3719295},
doi = {10.1145/3719295},
journal = {ACM Trans. Knowl. Discov. Data},
month = apr,
articleno = {80},
numpages = {27},
}

@ARTICLE{9357419,
  author={He, Tiantian and Bai, Lu and Ong, Yew-Soon},
  journal={IEEE Transactions on Cybernetics}, 
  title={Vicinal Vertex Allocation for Matrix Factorization in Networks}, 
  year={2022},
  volume={52},
  number={8},
  pages={8047-8060},
  doi={10.1109/TCYB.2021.3051606}}

@ARTICLE{8006248,
  author={Wu, Di and Luo, Xin and Wang, Guoyin and Shang, Mingsheng and Yuan, Ye and Yan, Huyong},
  journal={IEEE Transactions on Industrial Informatics}, 
  title={A Highly Accurate Framework for Self-Labeled Semisupervised Classification in Industrial Applications}, 
  year={2018},
  volume={14},
  number={3},
  pages={909-920},
  doi={10.1109/TII.2017.2737827}}

@ARTICLE{9159907,
  author={Wu, Di and Luo, Xin and Shang, Mingsheng and He, Yi and Wang, Guoyin and Wu, Xindong},
  journal={IEEE Transactions on Knowledge and Data Engineering}, 
  title={A Data-Characteristic-Aware Latent Factor Model for Web Services QoS Prediction}, 
  year={2022},
  volume={34},
  number={6},
  pages={2525-2538},
  doi={10.1109/TKDE.2020.3014302}}

@ARTICLE{9885025,
  author={Wu, Di and Luo, Xin and He, Yi and Zhou, Mengchu},
  journal={IEEE Transactions on Neural Networks and Learning Systems}, 
  title={A Prediction-Sampling-Based Multilayer-Structured Latent Factor Model for Accurate Representation to High-Dimensional and Sparse Data}, 
  year={2024},
  volume={35},
  number={3},
  pages={3845-3858},
  doi={10.1109/TNNLS.2022.3200009}}

\end{document}